\documentclass{article}

\usepackage{PRIMEarxiv}

\usepackage[utf8]{inputenc} 
\usepackage[T1]{fontenc}    
\usepackage{hyperref}       
\usepackage{url}            
\usepackage{booktabs}       
\usepackage{amsfonts}       
\usepackage{nicefrac}       
\usepackage{microtype}      
\usepackage{lipsum}
\usepackage{fancyhdr}       
\usepackage{graphicx}       
\graphicspath{{media/}}     

\title{MAX: Masked Autoencoder for X-ray Fluorescence in Geological Investigation
}

\author{
    An-Sheng Lee\\
    Department of Geosciences and Research Center for Future Earth\\
    National Taiwan University\\
    Taipei, Taiwan \\
    \texttt{ansheng@ntu.edu.tw} \\
    \And
    Yu-Wen Pao\\
    Department of Computer Science and Information Engineering \\
    Taipei, Taiwan \\
    \texttt{b09902016@csie.ntu.edu.tw} \\
    \And
    Hsuan-Tien Lin \\
    Department of Computer Science and Information Engineering \\
    Taipei, Taiwan \\
    \texttt{htlin@csie.ntu.edu.tw} \\
    \And
    Sofia Ya Hsuan Liou\\
    Department of Geosciences and Research Center for Future Earth\\
    National Taiwan University\\
    Taipei, Taiwan \\
    \texttt{yhliou@ntu.edu.tw} 
}

\begin{document}
\maketitle

\begin{abstract}
Pre-training foundation models has become the de-facto procedure for deep learning approaches, yet its application remains limited in the geological studies, where in needs of the model transferability to break the shackle of data scarcity.
Here we target on the X-ray fluorescence (XRF) scanning data, a standard high-resolution measurement in extensive scientific drilling projects.
We propose a scalable self-supervised learner, masked autoencoders on XRF spectra (MAX), to pre-train a foundation model covering geological records from multiple regions of the Pacific and Southern Ocean.
In pre-training, we find that masking a high proportion of the input spectrum (50\%) yields a nontrivial and meaningful self-supervisory task.
For downstream tasks, we select the quantification of XRF spectra into two costly geochemical measurements, CaCO$_3$ and total organic carbon, due to their importance in understanding the paleo-oceanic carbon system.
Our results show that MAX, requiring only one-third of the data, outperforms models without pre-training in terms of quantification accuracy.
Additionally, the model's generalizability improves by more than 60\% in zero-shot tests on new materials, with explainability further ensuring its robustness.
Thus, our approach offers a promising pathway to overcome data scarcity in geological discovery by leveraging the self-supervised foundation model and fast-acquired XRF scanning data. 
\end{abstract}


\pagebreak
\section{Introduction}
The properties of geological records, spanning chemical, and physical to biological, infer environmental variations through history and give substantial understanding to the Earth.
For instance, calcium carbonate (CaCO$_3$) and total organic carbon (TOC) are essential applications for reconstructing paleoceanographic carbon systems \cite{archer1994effect,seki2019high}.
These properties often require tedious work to acquire quantitative measurements.
Hence, the spatial/temporal coverage and resolution are limited, which induces high uncertainty in the research conclusion.
To address this limit, simulating these properties by quantifying the fast and high-resolution measurements has been proposed in studies.

X-ray fluorescence spectrometry (XRF) together with radiographic analysis and X-ray diffraction form the base of X-ray applications, which stand for three X-ray photon phenomena: fluorescence, adsorption, and scatter \cite{jenkins2000x}.
The XRF is the electromagnetic radiation produced by the falling of electrons from high to low energy states following the ejection of inner orbital electrons.
This emission can be aligned with known atomic numbers so the recorded spectrum allows the identification of elemental abundance.
The XRF core scanning technique is built upon this theory but without time-consuming sample pre-treatments.
It thus non-destructively measures the geological sample's semi-quantitative properties in a faster and higher resolution ($\geq 100 \mu m$) and becomes the standard procedure in miscellaneous geological studies.

Quantifying these XRF core scanning data to desired properties has majorly been studied in applying statistical analyses on XRF-derived elemental data \cite{croudace2006itrax}, \cite{Weltje2015}.
However, acquiring elemental data relies on experience and tailor-made fine-tuning of the analytic software, which is often overlooked due to the large amount of scanning.
Lee et al. (2022) \cite{lee2022quantifying} apply the conventional ML techniques to the raw XRF spectra to avoid this manual bias and include comprehensive scanning information.
Its quantification result offers an improved accuracy. However, the conventional ML techniques' scalability and capability for building general models covering vast material and task variation are constrained.
Most quantification approaches tend to train their project-scale models independently from scratch.
As a result, building accurate models usually requires enormous data for each project.
What's more, these methods are lack of generalizability across different tasks.

Recently, masked encoding opens a wide gate for self-supervised learning to pre-train foundation models on an enormous amount of unlabeled data.
A pre-training objective of predicting the 15\% masked-out input text sequence via a transformer-based model (BERT) is proposed \cite{devlin2019bertpretrainingdeepbidirectional} as a milestone.
This masking strategy in language has soon migrated to the computer vision community and the masking ratio is increased to 75\% to encourage the model to learn useful vision features \cite{he2021maskedautoencodersscalablevision}.
These foundation models demonstrate the superiority regarding performance and data demand in downstream tasks, and thus become the de-facto standard protocol.
The beauty of foundation model's transferability will be the key solution to the scarcity of geological data by uniting the contributions of dispersed projects.
Yet, there is neither off-the-shelf study in foundation models nor benchmark datasets for geological investigations.
A new chapter is ahead.

\begin{figure}
    \centering
    \includegraphics[width=1\linewidth]{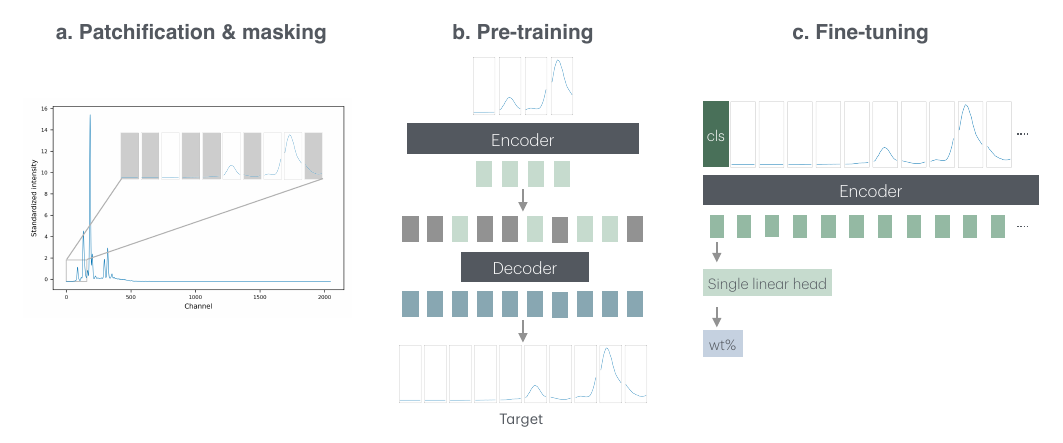}
    \caption{Scheme for three main components in MAX. \textbf{a} We split a spectrum into fixed-size (16 channels) patches and randomly masked out patches (labeled in gray). Only the first 10 patches are demonstrated here. The channel is the raw format, which will be converted into energy format by $20eV channel^{-1}$. \textbf{b} During pre-training, the leftover patches are fed to a ViT-base encoder for representation. The encoded patches (green) are then processed with the mask tokens (gray) by a small Transformer decoder that reconstructs the original spectrum in channels. \textbf{c} During fine-tuning, the encoder is applied to uncorrupted spectra supplemented with a [cls] token. The decoder is replaced by a single linear head taking only the [cls] token for regression tasks (i.e., quantifying CaCO$_3$ and TOC in wt\%).}
    \label{fig:scheme}
\end{figure}

We introduce a XRF representation model called MAX (Figure \ref{fig:scheme}), which stands for transformer-based masked autoencoder for XRF and is inspired by the masked image modeling (MAE \cite{he2021maskedautoencodersscalablevision}).
To overcome the data paucity, we collect measurements of geological records from the multiple international scientific drilling projects \cite{chao2022xdsa}, covering regions in the Pacific and Southern Ocean, and compile into ML-ready format for training.
In total, MAX is pre-trained by 55,211 XRF spectra. Each XRF spectrum is randomly masked out in a certain portion as the pre-training input for MAX to learn the reconstruction of missing values, which is a heuristic to learn XRF spectral characteristics.
Then, it is fine-tuned on the downstream tasks: quantifying XRF spectra into geochemical properties (CaCO$_3$ and TOC).
MAX proves its potential by requiring only 1/3 of the data for fine-tuning to outperform the baseline work \cite{lee2022quantifying}, which significantly reduces the burden of tedious laboratory measurements.
This study sets a milestone for filling the gap in foundation models and transcending the barriers of project-scale limitation.
With this pilot success, we are able to persuade more data owners to enrich the dataset and upscale the model from the current lightweight version, accordingly.

\section{Materials and methods}
\label{sec:methods}
Our method comprises two stages, the pre-training stage, and the fine-tuning stage, following the same design principle of the previous works from different domains like neural language processing \cite{devlin2019bertpretrainingdeepbidirectional} and computer vision \cite{he2021maskedautoencodersscalablevision}.
The pre-trained stage is to train a model that reconstructs the masked data by self-supervised learning.
In the fine-tuning stage, we use the part of the model weights from the previous stage to construct a regressor to predict the value for each downstream task by supervised learning.

The dataset is identical to the baseline (conventional ML approach \cite{lee2022quantifying}) for comparison.
The materials cover multiple regions at the high-latitude sectors of the Pacific and Southern Ocean are measured at the Alfred-Wegener-Institut Helmholtz-Zentrum für Polar- und Meeresforschung, Bremerhaven, Germany.
This dataset has 59,828 XRF spectra, 2,254 CaCO$_3$ measurements, and 2,363 TOC measurements.
The two geochemical properties are aligned with the XRF data based on depth. The details of the cores and measurements, please refer to the baseline \cite{lee2022quantifying}.
We select the data from three cores (cores PS75-056-1, LV28-44-3-n, and SO264-69-2) as the case study, staying untouched until the evaluation.
The data from the rest of the cores are randomly separated into training and validation sets with a 4:1 ratio.

\paragraph{Data availability}
We adopt the XRF spectra, CaCO$_3$, and TOC data from Chao et al. (2022) \cite{chao2022bcma} for fine-tuning.
The XRF spectra for pre-training are the unpublished raw data of XRF scanning results in Chao et al. (2022) \cite{chao2022xdsa}.
The data are prepared and split into ML-ready datasets, which are open to the public at \url{https://huggingface.co/datasets/paoyw/max-dataset/tree/main}.
The analyzing results are included in the supplement \href{https://docs.google.com/spreadsheets/d/1RuJPA2AN4f5MaxOf2EG8_FsKz8nfvEKb/edit?usp=sharing&ouid=106531072920562885700&rtpof=true&sd=true} {excel file} or can be reproduced through the open code.

\paragraph{Code availability}
The custom code can be found at the GitHub  \url{https://github.com/dispink/xpt}, which is open to the public.

\subsection{Pre-training stage: MAX for reconstruction}
The model structure of the masked autoencoder for X-ray fluorescence (MAX) is mainly based on MAE \cite{he2021maskedautoencodersscalablevision}, an autoencoder for images.
As with all of the autoencoders, MAX comprised two parts, the encoder and the decoder.
The encoder encodes the input data into the latent representation, and the decoder reconstructs the input data from the latent representation.
However, the autoencoders from the computer vision domain cannot naively be applied to the XRF data.
Compared with the image data, the XRF is different in the range of the value and the shape of the data.

The value range of the XRF data is unbounded and commonly from $0$ to more than $65000$ dramatically with different channels in the dataset, and that of the image data is bounded in the range of $[0, 1]$.
To deal with the unbounded range and the large differences among the channel and the instance, a proper data transformation to scale is needed.
We experimented with three kinds of transformation, instance-wised normalization, channel-wised normalization, and logarithm transformation.

\pagebreak
The instance-wised normalization calculates the mean and standard deviation for every instance (i.e., spectrum) to normalize the values.
\begin{equation}
\textit{instance-wised-normalization}(X) = (\frac{x_1 - \mu}{\sigma}, \dots, \frac{x_n - \mu}{\sigma})
\end{equation}

\begin{equation}
\mu = \sum_{i=1}^n\frac{x_i}{n}, 
\sigma = \sqrt{\frac{1}{n}\sum_{i=1}^n(x_i - \mu)}
\end{equation}

where $X = (x_1, ..., x_n)$ is a data point, $x_i$ is the value at $i$th channels, $n$ is the total number of the channels for a data point. This normalization makes the signal independent of tube ageing or any other factor affecting the primary beam intensity.

The channel-wised normalization normalizes the values along each channel for all data points.
\begin{equation}
\textit{channel-wised-normalization}(X) = (\frac{x_1 - \mu_1}{\sigma_1}, \dots, \frac{x_n - \mu_n}{\sigma_n})
\end{equation}

\begin{equation}
\mu_i = \sum_{X\in D_{train}} \frac{x_i}{|D_{train}|}, 
\sigma_i = \sqrt{\frac{1}{|D_{train}|}\sum_{X\in D_{train}}(x_i - \mu_i)}
\end{equation}

where $X = (x_1, ..., x_n)$ is a data point, $x_i$ is the value at $i$th channels, $n$ is the total number of the channels for a data point, and $D_{train}$ is the training set. $\mu_i$ and $\sigma_i$ is the mean and the standard deviation computes along $i$th channel from the training set.  However, the standard deviation may be $0$, because all the values at one channel is zero in the training set. We deal with this by division by $1$ instead.
The logarithm transformation is following. 
\begin{equation}
\textit{logarithm-transformation}(X) = (\ln{(x_1 + 1)}, \dots, \ln{(x_n + 1)})
\end{equation}

where $X = (x_1, ..., x_n)$ is a data point.
We add $1$ before logarithm to avoid the zero value.
Based on the experiment results from the \hyperref[sec:exp-data-transformation]{section}, we choose the instance-wised normalization as the default setting.

The shape of the XRF data is 1 dimensional with the channel only, and that of the image data is 3 dimensional with the width, the height, and the color channel.
Hence, the patchfication, which means dividing an image into regular non-overlapping patches in ViT \cite{dosovitskiy2020image} and MAE \cite{he2021maskedautoencodersscalablevision}, needs to be modified.
Instead of patchifying the data into small rectangles in computer vision, we patchify the 1D XRF data into segments.
For each segment, we choose 16 non-overlapping and consecutive channels out of the $2048$ channels of the XRF data.
After the patchification, we apply the 1D convolution to map the $16$ to the designed number of  channels, which is $1024$ in our case.
The shape of a batch of input data becomes \textit{(the batch size, the number of channels,
and the number of segments)}, which is the same shape as the input of the encoder in ViT \cite{dosovitskiy2020image}. 
The patchification also gives MAX the flexibility for different lengths of XRF spectra from different scanning machine series. This is vital in the future when enlarging data collection. 

For the remaining parts of the encoder and the decoder, we apply the same design in MAE \cite{he2021maskedautoencodersscalablevision}.
The encoder first adds the data with positional embedding, then applies a series of Transformer blocks \cite{Vaswan2017} on the unmasked parts, and outputs the encoded patches.
The inputs of the decoder are the encoded patches from the encoder and the shared mask token for the masking part.
The decoder first adds the positional embeddings to the inputs and then applies another series of Transformer blocks.
After that, it projects the outputs back to the numbers of the channels of the patchified data by a linear layer and then transforms the shape of the output back to that of the original data.

The target of MAX at the pre-training stage is to reconstruct the XRF data.
The loss function is the mean square error loss between the original input and the reconstructed output.
Following BERT \cite{devlin2019bertpretrainingdeepbidirectional} and MAE \cite{he2021maskedautoencodersscalablevision}, we calculate the loss on masking parts only.

For training, we use AdamW \cite{loshchilov2017decoupled} as the optimizer and the linear-warm up for $10$ epochs followed by the cosine annealing as the learning rate scheduler.
We pre-train the model using the entire training set with $100$ epochs, $256$ batch size, and $1e-4$ learning rate.
Based on the discussion and experiment in the next \hyperref[sec:exp-mask-ratio]{section}, we choose $0.5$ as our default masking ratio.

\pagebreak
\subsection{Fine-tuning stage: MAX for regression}
Since the values to be predicted vary a lot in downstream tasks (CaCO$_3$ and TOC), we normalize them by the mean and the standard deviation of the training set first.
For evaluation, we will transform them back to the original scale.

At the fine-tuning stage, we adopt the well-trained encoder from the pre-training stage.
After adding the positional embedding to the projected and patchified data, we append the classification token to it.
After passing through the transformer blocks, we take out the latent representation responding to the classification token and pass it to a linear layer with only one output channel to predict the value.
The loss function is the mean square error between the prediction and the ground truth.
Instead of freezing parts of the pre-trained model weights or linear probing \cite{he2021maskedautoencodersscalablevision}, we optimize the model weights entirely.
We choose the AdamW \cite{loshchilov2017decoupled} as our fine-tuning optimizer, and the linear warm-up for $10\%$ of the epochs followed by the cosine annealing as the learning rate scheduler.
We fine-tune the model using the entire training set for each downstream task with $100$ epochs, $256$ batch size, and $1e-5$ learning rate.

\subsection{Case study}
We first apply the MAX fine-tuned in the previous stage on the entire case study data to obtain the zero-shot accuracy. Then, the case study is randomly split into the training and validation sets with 4:1 ratio. We fine-tune the pre-trained MAX model (without previous fine-tuning) using different amount of data in the case study for each downstream task with $256$ batch size. The epoch and learning rate are tuned for each data amount setting.

\subsection{Saliency map}
To study the explainability of our model, we use the saliency map \cite{simonyan2014deepinsideconvolutionalnetworks} of the model as the main method.
The authors started from the linear classification model for the classification class $c$:
\begin{equation}
    S_c(I) = w_c^TI+ b_c
\end{equation}
where the $S_c$ is the class score function for the class $c$, $I$ is the input image, $w_c$ and $b_c$ are the weight vector and the bias of the model respectively.
The magnitude of elements of $w$ indicates the importance of the corresponding pixels of $I$ for the target class $c$.
For the non-linear models like ViT \cite{dosovitskiy2020image}, we can approximate them by the first-order Taylor expansion:
\begin{equation}
S_c(I) \approx w^TI+ b
\end{equation}
where $w$ is the derivative of $S_c$ with respect to the image $I$ at the point $I_0$:
\begin{equation}
w = \frac{\partial S_c}{\partial I} \Bigr|_{I_0}
\end{equation}
One of the interpretations of this method is that the magnitude of the derivative shows which channels to be changed the least to affect the most of the loss value.
In our experiment, we calculate the gradient by the mean square error between the prediction and the ground truth value.
The saliency map $M$ can be formulated as:
\begin{equation}
M = \Bigr|\frac{\partial MSE(y, \hat{y})}{\partial X} \Bigr|_{X_0}
\end{equation}

where $M$ is the saliency map, $y$ is the ground truth value, and $\hat{y}$ is the prediction of the model with $X_0$ as the input.

\section{Experiment Results}\label{sec:experiment-results}
\subsection{Mask ratio}\label{sec:exp-mask-ratio}
\begin{figure}
    \centering
    \includegraphics[width=0.55\linewidth]{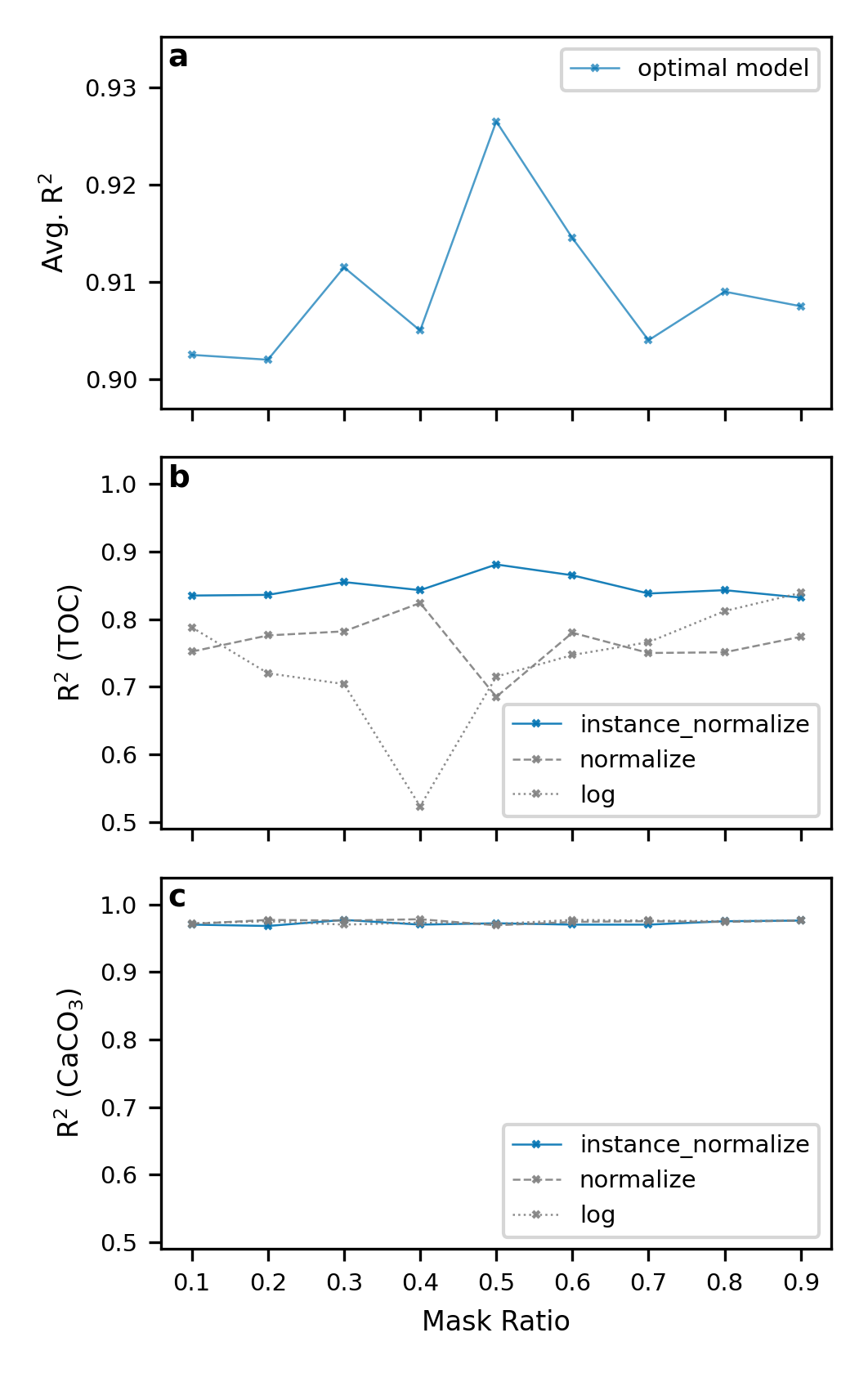}
    \caption{MAX accuracy v.s. mask ratio in pre-training. \textbf{a} The optimal accuracy (averaged R$^2$) in each mask ratio setting.\textbf{ b, c} The accuracy (R$^2$) of each transformation and mask ratio settings in tasks TOC and CaCO$_3$.}
    \label{fig:r2_vs_ratio}
\end{figure}

Figure \ref{fig:r2_vs_ratio}a displays the influence of mask ratio during pre-training.
The ratio optimizing MAX performance on the downstream task is located at 0.5.
This high mask ratio implies that the masked patches of a spectrum can be easily interpolated with neighboring patches if the mask ratio is low.
In this case, the information density of an XRF spectrum is low and falls closer to the vision (mask ratio: 0.75) \cite{he2021maskedautoencodersscalablevision} than to the language (mask ratio: 0.15) \cite{devlin2019bertpretrainingdeepbidirectional}.
For comparison, masking out a few words in a sentence could produce great challenges for speculating the original sentence. 

\pagebreak

\subsection{Data transformation}\label{sec:exp-data-transformation}
The initial weights in ML models are usually centered at 0\cite{glorot2010understanding, he2015delving}.
If the data have large or extreme values, it requires a good recipe for manipulations, like large initial learning rates.
In our pilot study, it requires a $1e-3$ learning rate to get the model barely converged.
Thus, data transformation is recommended for ML to keep things simple.
As shown in the figure \ref{fig:r2_vs_ratio}, the model performance is dominantly affected by the TOC accuracy.
The instance-wised normalization overall gives the best accuracy, which may inform the severe bias caused by tube aging or any other factor affecting the primary beam intensity.
This is reasonable due to the long scanning period required for this large amount of cores.
Furthermore, instance-wise data transformation is commonly adopted in XRF core scanning studies, such as ratio \cite{LOWEMARK20111250} and log-ratio \cite{WELTJE2008423, LEE201944} normalization.

\subsection{Spectral reconstruction}\label{sec:spectral-reconstruction}
\begin{figure}
    \centering
    \includegraphics[width=1\linewidth]{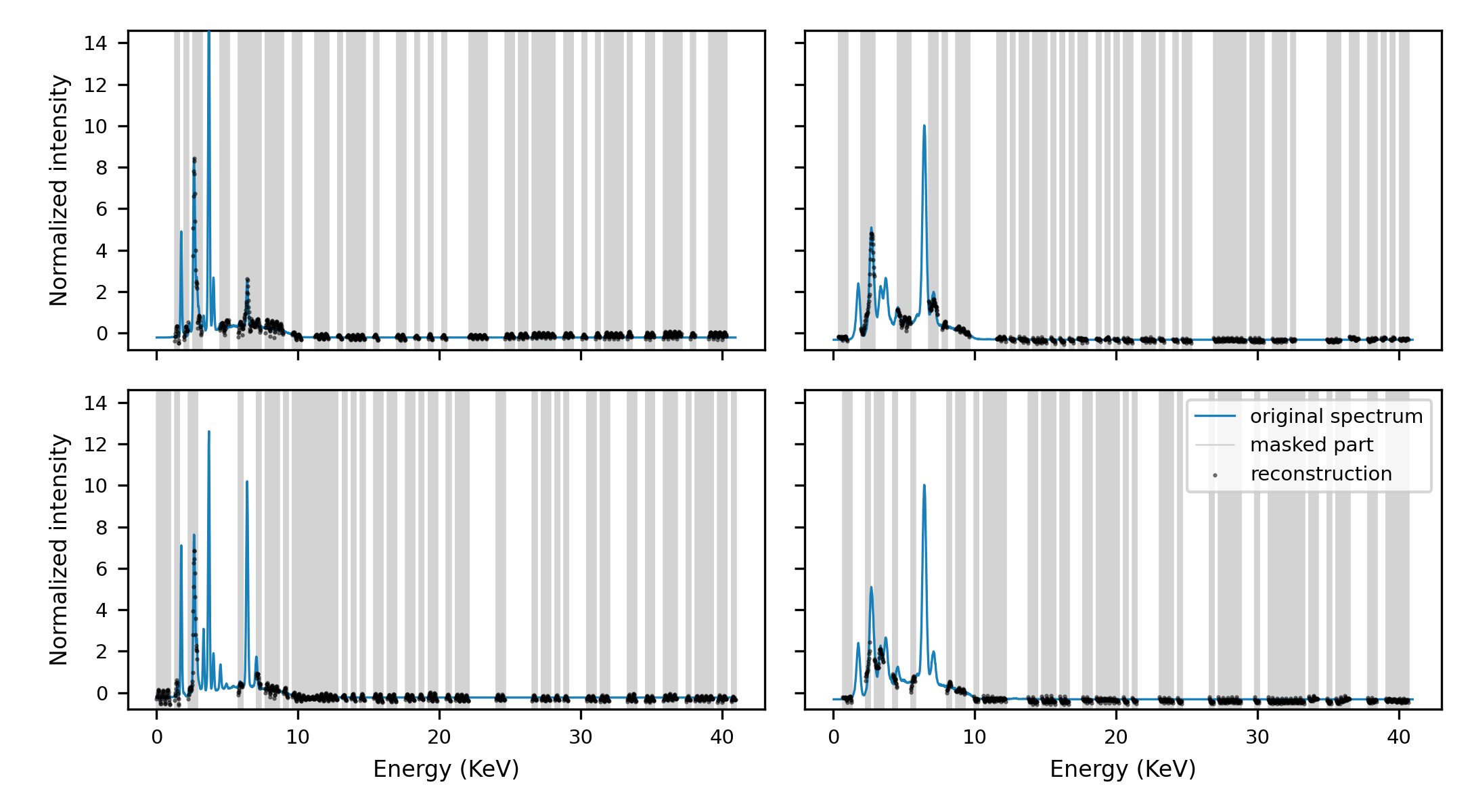}
    \caption{Example results on validation spectra. For each figure, we show the original spectrum (blue line), the masked part (gray intervals), and our MAX reconstruction (black dots). The masking ratio is $0.5$, leaving only 64 out of 128 patches. }
    \label{fig:reconstruction}
\end{figure}

After we found the optimal pre-training setting based on the downstream tasks' performance, it is essential to confirm our concept of self-supervised pre-training: whether MAX can reconstruct the masked spectrum.
Figure \ref{fig:reconstruction} shows some examples of successful spectral reconstruction even under the heavy mask ratio.
The R$^2$ in the validation set is 0.973. The migration of MAE \cite{he2021maskedautoencodersscalablevision} from computer vision to XRF spectrometry has been proven to be feasible.

\subsection{Data amount for training}\label{sec:exp-data-amount}
\begin{figure}
    \centering
    \includegraphics[width=0.55\linewidth]{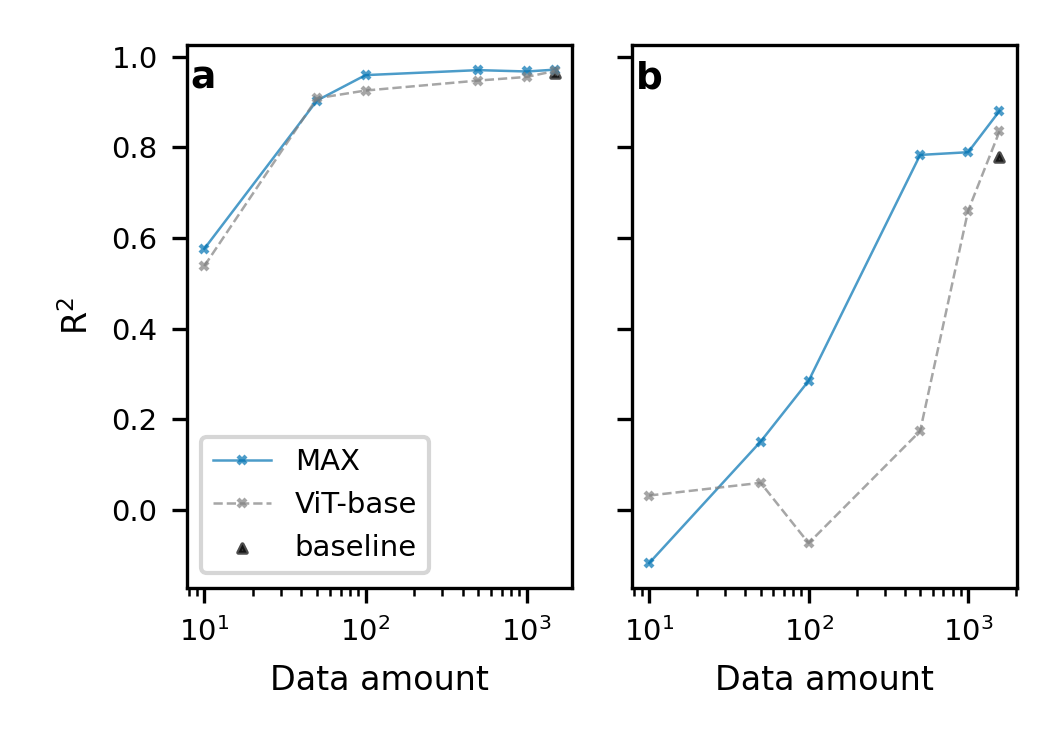}
    \caption{Validation accuracy v.s. fine-tuning data amount. Three series of models (MAX,  train-from-scratch ViT-base, and baseline \cite{lee2022quantifying}) are compared in tasks CaCO$_3$ (\textbf{a}) and TOC (\textbf{b}). }
    \label{fig:r2_vs_data}
\end{figure}

We fine-tune MAX with different amounts of data for downstream tasks to see how much data MAX needs to get decent performance.
Then, we compare their performance with the train-from-scratch ViT-base model \cite{dosovitskiy2020image} (relevant to MAX without pre-training) and the baseline (conventional ML approach \cite{lee2022quantifying}) on the validation set.
Figure \ref{fig:r2_vs_data} shows that MAX performance reaches a plateau starting fine-tuned by 50 and 500 data points for tasks CaCO$_3$ and TOC, respectively.
Compared to the eminent CaCO$_3$ quantification, MAX's slower growth rate and lower value of TOC accuracy (defined by R$^2$) point out the difficulty of TOC quantification, which is due to the lack of XRF signal to organic elements (i.e., H, C, N, O) \cite{richter2006avaatech}.
Nevertheless, MAX surpasses the baseline accuracy for both tasks after only being fine-tuned by 500 data points, which is merely 1/3 of the whole training dataset (CaCO$_3$: 1488; TOC: 1573).
In the CaCO$_3$ task, MAX provides comparable accuracy to the baseline \cite{lee2022quantifying} when using only 50 data points for fine-tuning.
MAX also outperforms the train-from-scratch ViT-base model in a few training data scenarios, except for the extreme case (10 data points) in the TOC task.
When the training data amount approaches to the whole training data set, the superiority of MAX reduces.
This phenomenon suggests the main benefit of pre-training: reducing training data, and also the high capacity of ViT-base. 

\subsection{Case study}
We utilize the fine-tuned MAX on a wild test as a case study: three new cores across the Pacific and Southern Ocean.
Table \ref{tab:r2_test} indicates our MAX has 0.365 and 0.978 improvements in R$^2$ values of CaCO$_3$ and TOC compared to the baseline \cite{lee2022quantifying}.
However, the performance on TOC is still not satisfying.
It may be explained by the noticeable difference in TOC data distribution between the training set (mean: 0.21, std: 0.18) and the case study (mean: 0.38, std: 0.26).
The mean value in the case study is nearly twice the training set.
This extreme extrapolation is beyond our TOC model's ability.
The CaCO$_3$ data remain similar so the performance stays excellent.
We, therefore, encourage future applications to check their data distribution before applying our fine-tuned model directly.

\begin{table}
    \caption{Zero-shot accuracy (R$^2$) of MAX and the baseline \cite{lee2022quantifying} in the case study.}
    \label{tab:r2_test}
    \centering
    \begin{tabular}{ccc}
        \toprule
        R$^2$ &MAX &Baseline* \cite{lee2022quantifying}\\
        \midrule
        CaCO$_3$ &0.975&  0.61\\ 
        TOC &0.018&  -0.96\\
        \bottomrule
    \end{tabular}\\
    \footnotesize{*The baseline's R$^2$ values were miscalculated with only cores LV28-44-3-n and SO264-69-2, leaving out core PS75-056-1. Hence, we treat the values as the lower bound of error.}
\end{table}

\begin{figure}
    \centering
    \includegraphics[width=0.65\linewidth]{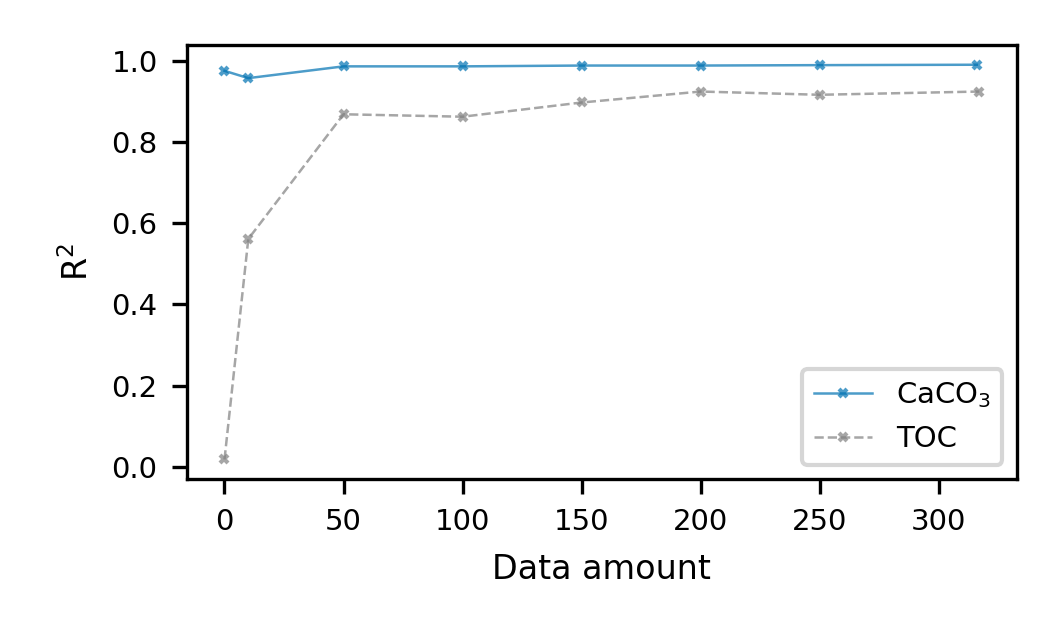}
    \caption{Accuracy v.s. fine-tuning data amount in case study. The accuracy (R$^2$) is evaluated in the validation set of the case study, except for the zero-shot test. The zero-shot accuracy, marked at 0 data amount, is evaluated by the entire case study data.} 
    \label{fig:case_data}
\end{figure}

If the data are different, like TOC in the case study, the pre-training benefit of MAX comes into effect.
By fine-tuning MAX with merely 13\% (N: 50) of the entire data set (N: 394), TOC reaches a stable and good accuracy ({R$^2$}: 0.868, Figure \ref{fig:case_data}).

\section{Model explainability}\label{sec:model-explainability}
\begin{figure}[h]
    \centering
    \includegraphics[width=1\linewidth]{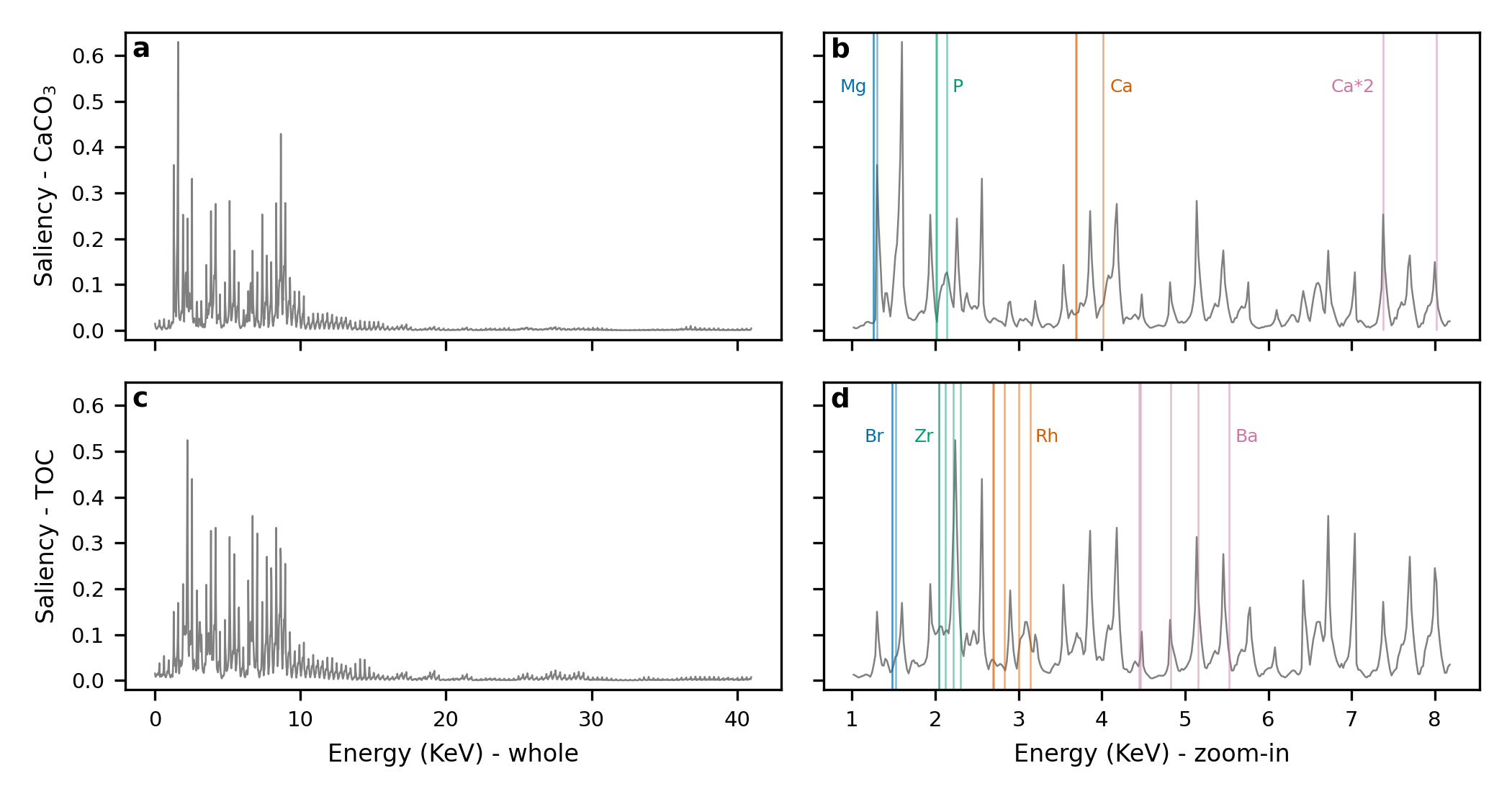}
    \caption{Saliency maps of CaCO$_3$ and TOC models. \textbf{a, c} The whole map. \textbf{b, d} The zoom-in map with known elemental emission lines.}
    \label{fig:saliency}
\end{figure}

Besides the MAX performance, we explore the aspects of spectral appearance, captured inside models.
The often-asked question, "Which part of the spectrum contributes more to the model's prediction?" is answered by the averaged saliency map \cite{simonyan2014deepinsideconvolutionalnetworks} of our downstream models from one batch of data.
Figure \ref{fig:saliency}a and c show that high saliency concentrates in 1 -10 KeV. We find the CaCO$_3$ model taking Mg and P as important contributing elements (Figure \ref{fig:saliency}b).
Magnesium commonly co-exists with Ca in carbonates due to their abundance in seawater and similar chemical properties.
Phosphorous plays an important role in oceanic biogenic productivity \cite{RUTTENBERG20012149}, which corresponds to the formation of CaCO$_3$.
The original Ca emission lines do not match any saliency peak.
However, we can find matches at twice the original Ca energy, which implies the sum peaks effect.
Two Ca photons hit the detector almost at the same time, which is possible in spectra that have a high concentration of Ca.
The peaks also match with the energy of rare elements, like Nb, Ru, Ce, Nd, Sm, Ho, W, Re, and Zn. We prefer to interpret them as artifacts without further confirmation. 

The TOC model (Figure \ref{fig:saliency}d) has saliency peaks matching with the energy of Zr, Ba, Rh, and other rare elements, which we interpret as artifacts (Dy, Ru, Ho, W, Zn).
Zirconium usually is rich in coarse sediments due to its composition of resistant minerals \cite{soton380321}.
It is a good contrasting proxy to organic matters, which are often fine-grains and lack of minerals.
Barium is a vital proxy for paleoproductivity because of its occurrence within skeletal and organic detritus \cite{cp-6-63-2010}.
Two of the Rh emission lines (2.83 and 3.00 Ke) match the saliency peaks with slight energy loss.
This could indicate that the model is taking the inelastic scattering of the X-ray tube source (Rh) photons into account, which is also called Compton peaks.
This artifact is often seen in the low-density samples and thus treated as an indirect TOC proxy \cite{croudace2006itrax}.
The commonly used marine TOC proxy, Br, doesn't match any saliency peaks.
We speculate this proxy is not effective due to the samples' low concentration of TOC.
In the end, the model rather depends on other parts of the spectrum to jointly calibrate this difficult task.

The productivity proxies, P and Ba, can be found salient both in the two models, but with slight differences.
This phenomenon could imply the varying positive correlation of biogenic activities to CaCO$_3$ and TOC.
In addition, there are still many saliency peaks that remain unexplained, which is worthy of further investigation.

\pagebreak
\section{Conclusion}
MAX successfully migrates the success in the natural language (BERT \cite{devlin2019bertpretrainingdeepbidirectional}) and vision (MAE \cite{he2021maskedautoencodersscalablevision}) through the perspective of XRF spectrometry to the geological field.
A transformer-based deep model is now able to self-learn XRF spectral characteristics through masked spectrum modeling.
This pre-training process provides not only strong regularization to obtain good performance but also great efficiency in computation.
Its scalability is promoted by the benefit of the transformer and the simplicity of self-supervised learning.
We hypothesize that the outstanding performance of MAX (higher accuracy and fewer training data needed in the downstream tasks) is supported by the rich hidden representation of the XRF spectrum learned during pre-training.
As a foundation model, MAX is ready to transfer the representation for broad geological tasks.
Moreover, it is potentially transferable to other X-ray applications, extending achievements beyond the XRF.

MAX opens a new chapter of geological investigation. More studies shall be launched next.
First, the scale of data requires a significant increase.
We believe our training data (55K) is still not enough for the ViT-base model, the backbone of MAX, compared to its original research \cite{dosovitskiy2020image}.
Furthermore, there are still ViT-large and ViT-huge out there for scaling up model size and data simultaneously.
If we look at the leading foundation models (e.g., Meta Llama \cite{touvron2023llamaopenefficientfoundation}, Google PaLM \cite{chowdhery2023palm}, Microsoft BEiT \cite{wang2022imageforeignlanguagebeit}), which have billions or trillions of model sizes and data, MAX is minuscule.
Besides the size, the quality of data needs to be improved.
The imbalanced spectra from different materials should be carefully balanced and cleaned, which is neglected in this work.
Once done, more deep learning techniques are worthy for exploration to increase the efficiency of model application, such as untargeted domain adaptation\cite{wang2018deep, fang2024source}, few-shot learning \cite{song2023comprehensive}, prompting \cite{liu2023pre, sahoo2024systematic}, and parameter-efficient fine-tuning \cite{houlsby2019parameter, lester2021power, hu2021lora, han2024parameter}.
Moreover, the scientific community shows a strong interest in model explainability. What we offer is merely an appetizer.
It could be advanced by tailor-made data and experiment design, or other model architectures (e.g., Physics-informed neural networks \cite{RAISSI2019686}).
After all, we hope MAX will inspire future works to enrich this new chapter.

\section*{Author contributions statement}
A.S.L. and Y.W.P.: conceptualization, methodology, investigation, data curation, writing—original draft. \\
H.T.L. and S.Y.H.L.: supervision, resources, funding acquisition, writing—review and editing. 

\section*{Acknowledgments}
We thank the crew and the science parties of different cruises for their contributions to core and sample acquisition on the respective expeditions.
We are very grateful to Dr. Weng‐Si Chao, Dr. Lester Lembke‐Jene, and Dr. Frank Lamy for providing these data.
We acknowledges financial supports from 
(1) the National Science and Technology Council of Taiwan (project number:  NSTC 112-2116-M-002-009, NSTC 113-2628-E-002-00), 
(2) the National Taiwan University Research Center for Future Earth from the Featured Areas Research Center Program within the Higher Education Sprout Project framework funded by the Ministry of Education of Taiwan, and
(3) the National Taiwan University Center for Data Intelligence via NTU-113L900901.

\bibliographystyle{unsrt}  
\bibliography{references}

\begin{thebibliography}{10}

\bibitem{archer1994effect}
D~Archer and Ernst Maier-Reimer.
\newblock Effect of deep-sea sedimentary calcite preservation on atmospheric
  co2 concentration.
\newblock {\em Nature}, 367(6460):260--263, 1994.

\bibitem{seki2019high}
Arisa Seki, Ryuji Tada, Shunsuke Kurokawa, and Masafumi Murayama.
\newblock High-resolution quaternary record of marine organic carbon content in
  the hemipelagic sediments of the japan sea from bromine counts measured by
  xrf core scanner.
\newblock {\em Progress in Earth and Planetary Science}, 6(1):1--12, 2019.

\bibitem{jenkins2000x}
Ron Jenkins.
\newblock X-ray techniques: overview.
\newblock {\em Encyclopedia of analytical chemistry}, pages 1--20, 2000.

\bibitem{croudace2006itrax}
Ian~W Croudace, Anders Rindby, and R~Guy Rothwell.
\newblock Itrax: description and evaluation of a new multi-function x-ray core
  scanner.
\newblock {\em Geological Society, London, Special Publications},
  267(1):51--63, 2006.

\bibitem{Weltje2015}
G.~J. Weltje, M.~R. Bloemsma, R.~Tjallingii, D.~Heslop, U.~R{\"o}hl, and Ian~W.
  Croudace.
\newblock {\em Prediction of Geochemical Composition from XRF Core Scanner
  Data: A New Multivariate Approach Including Automatic Selection of
  Calibration Samples and Quantification of Uncertainties}, pages 507--534.
\newblock Springer Netherlands, Dordrecht, 2015.

\bibitem{lee2022quantifying}
An-Sheng Lee, Weng-Si Chao, Sofia Ya~Hsuan Liou, Ralf Tiedemann, Bernd
  Zolitschka, and Lester Lembke-Jene.
\newblock Quantifying calcium carbonate and organic carbon content in marine
  sediments from xrf-scanning spectra with a machine learning approach.
\newblock {\em Scientific Reports}, 12(1):20860, 2022.

\bibitem{devlin2019bertpretrainingdeepbidirectional}
Jacob Devlin, Ming-Wei Chang, Kenton Lee, and Kristina Toutanova.
\newblock Bert: Pre-training of deep bidirectional transformers for language
  understanding, 2019.

\bibitem{he2021maskedautoencodersscalablevision}
Kaiming He, Xinlei Chen, Saining Xie, Yanghao Li, Piotr Dollár, and Ross
  Girshick.
\newblock Masked autoencoders are scalable vision learners, 2021.

\bibitem{chao2022xdsa}
Weng si~{Chao}, An-Sheng {Lee}, Ralf {Tiedemann}, Lester {Lembke-Jene}, and
  Frank {Lamy}.
\newblock {XRF down-core scanning and bulk chemistry measurements of sediments
  from the high latitude sectors of Pacific Ocean}, 2022.

\bibitem{chao2022bcma}
Weng-Si Chao, An-Sheng Lee, Ralf Tiedemann, Lester Lembke-Jene, and Frank Lamy.
\newblock Bulk chemistry measurements and xrf spectra of sediments from the
  high latitude sectors of pacific ocean.
\newblock In {\em XRF down-core scanning and bulk chemistry measurements of
  sediments from the high latitude sectors of Pacific Ocean}. PANGAEA, 2022.

\bibitem{dosovitskiy2020image}
Alexey Dosovitskiy, Lucas Beyer, Alexander Kolesnikov, Dirk Weissenborn,
  Xiaohua Zhai, Thomas Unterthiner, Mostafa Dehghani, Matthias Minderer, Georg
  Heigold, Sylvain Gelly, et~al.
\newblock An image is worth 16x16 words: Transformers for image recognition at
  scale.
\newblock {\em arXiv preprint arXiv:2010.11929}, 2020.

\bibitem{Vaswan2017}
Ashish Vaswani, Noam Shazeer, Niki Parmar, Jakob Uszkoreit, Llion Jones,
  Aidan~N Gomez, \L~ukasz Kaiser, and Illia Polosukhin.
\newblock Attention is all you need.
\newblock In I.~Guyon, U.~Von Luxburg, S.~Bengio, H.~Wallach, R.~Fergus,
  S.~Vishwanathan, and R.~Garnett, editors, {\em Advances in Neural Information
  Processing Systems}, volume~30. Curran Associates, Inc., 2017.

\bibitem{loshchilov2017decoupled}
Ilya Loshchilov and Frank Hutter.
\newblock Decoupled weight decay regularization.
\newblock {\em arXiv preprint arXiv:1711.05101}, 2017.

\bibitem{simonyan2014deepinsideconvolutionalnetworks}
Karen Simonyan, Andrea Vedaldi, and Andrew Zisserman.
\newblock Deep inside convolutional networks: Visualising image classification
  models and saliency maps, 2014.

\bibitem{glorot2010understanding}
Xavier Glorot and Yoshua Bengio.
\newblock Understanding the difficulty of training deep feedforward neural
  networks.
\newblock In {\em Proceedings of the thirteenth international conference on
  artificial intelligence and statistics}, pages 249--256. JMLR Workshop and
  Conference Proceedings, 2010.

\bibitem{he2015delving}
Kaiming He, Xiangyu Zhang, Shaoqing Ren, and Jian Sun.
\newblock Delving deep into rectifiers: Surpassing human-level performance on
  imagenet classification.
\newblock In {\em Proceedings of the IEEE international conference on computer
  vision}, pages 1026--1034, 2015.

\bibitem{LOWEMARK20111250}
L.~Löwemark, H.-F. Chen, T.-N. Yang, M.~Kylander, E.-F. Yu, Y.-W. Hsu, T.-Q.
  Lee, S.-R. Song, and S.~Jarvis.
\newblock Normalizing xrf-scanner data: A cautionary note on the interpretation
  of high-resolution records from organic-rich lakes.
\newblock {\em Journal of Asian Earth Sciences}, 40(6):1250--1256, 2011.
\newblock Quaternary Paleoclimate of the Western Pacific and East Asia: State
  of the Art and New Discovery.

\bibitem{WELTJE2008423}
Gert~Jan Weltje and Rik Tjallingii.
\newblock Calibration of xrf core scanners for quantitative geochemical logging
  of sediment cores: Theory and application.
\newblock {\em Earth and Planetary Science Letters}, 274(3):423--438, 2008.

\bibitem{LEE201944}
An-Sheng Lee, Jyh-Jaan~Steven Huang, George Burr, Li~Cheng Kao, Kuo-Yen Wei,
  and Sofia Ya~Hsuan Liou.
\newblock High resolution record of heavy metals from estuary sediments of
  nankan river (taiwan) assessed by rigorous multivariate statistical analysis.
\newblock {\em Quaternary International}, 527:44--51, 2019.

\bibitem{richter2006avaatech}
Thomas~O Richter, Sjerry Van~der Gaast, Bob Koster, Aad Vaars, Rineke Gieles,
  Henko~C de~Stigter, Henk De~Haas, and Tjeerd~CE van Weering.
\newblock The avaatech xrf core scanner: technical description and applications
  to ne atlantic sediments.
\newblock {\em Geological Society, London, Special Publications},
  267(1):39--50, 2006.

\bibitem{RUTTENBERG20012149}
K.C. Ruttenberg.
\newblock Phosphorus cycle.
\newblock In John~H. Steele, editor, {\em Encyclopedia of Ocean Sciences},
  pages 2149--2162. Academic Press, Oxford, 2001.

\bibitem{soton380321}
R.G. Rothwell and I.W. Croudace.
\newblock Twenty years of xrf core scanning marine sediments: What do
  geochemical proxies tell us?
\newblock In W.~Croudace I and Rothwell R.G, editors, {\em Micro-XRF Studies of
  Sediment Cores: Applications of a non-destructive tool for the environmental
  sciences}, 17, pages 25--102. Springer, July 2015.

\bibitem{cp-6-63-2010}
M.~Ziegler, L.~J. Lourens, E.~Tuenter, and G.-J. Reichart.
\newblock High arabian sea productivity conditions during mis 13 \&ndash; odd
  monsoon event or intensified overturning circulation at the end of the
  mid-pleistocene transition?
\newblock {\em Climate of the Past}, 6(1):63--76, 2010.

\bibitem{touvron2023llamaopenefficientfoundation}
Hugo Touvron, Thibaut Lavril, Gautier Izacard, Xavier Martinet, Marie-Anne
  Lachaux, Timothée Lacroix, Baptiste Rozière, Naman Goyal, Eric Hambro,
  Faisal Azhar, Aurelien Rodriguez, Armand Joulin, Edouard Grave, and Guillaume
  Lample.
\newblock Llama: Open and efficient foundation language models, 2023.

\bibitem{chowdhery2023palm}
Aakanksha Chowdhery, Sharan Narang, Jacob Devlin, Maarten Bosma, Gaurav Mishra,
  Adam Roberts, Paul Barham, Hyung~Won Chung, Charles Sutton, Sebastian
  Gehrmann, et~al.
\newblock Palm: Scaling language modeling with pathways.
\newblock {\em Journal of Machine Learning Research}, 24(240):1--113, 2023.

\bibitem{wang2022imageforeignlanguagebeit}
Wenhui Wang, Hangbo Bao, Li~Dong, Johan Bjorck, Zhiliang Peng, Qiang Liu, Kriti
  Aggarwal, Owais~Khan Mohammed, Saksham Singhal, Subhojit Som, and Furu Wei.
\newblock Image as a foreign language: Beit pretraining for all vision and
  vision-language tasks, 2022.

\bibitem{wang2018deep}
Mei Wang and Weihong Deng.
\newblock Deep visual domain adaptation: A survey.
\newblock {\em Neurocomputing}, 312:135--153, 2018.

\bibitem{fang2024source}
Yuqi Fang, Pew-Thian Yap, Weili Lin, Hongtu Zhu, and Mingxia Liu.
\newblock Source-free unsupervised domain adaptation: A survey.
\newblock {\em Neural Networks}, page 106230, 2024.

\bibitem{song2023comprehensive}
Yisheng Song, Ting Wang, Puyu Cai, Subrota~K Mondal, and Jyoti~Prakash Sahoo.
\newblock A comprehensive survey of few-shot learning: Evolution, applications,
  challenges, and opportunities.
\newblock {\em ACM Computing Surveys}, 55(13s):1--40, 2023.

\bibitem{liu2023pre}
Pengfei Liu, Weizhe Yuan, Jinlan Fu, Zhengbao Jiang, Hiroaki Hayashi, and
  Graham Neubig.
\newblock Pre-train, prompt, and predict: A systematic survey of prompting
  methods in natural language processing.
\newblock {\em ACM Computing Surveys}, 55(9):1--35, 2023.

\bibitem{sahoo2024systematic}
Pranab Sahoo, Ayush~Kumar Singh, Sriparna Saha, Vinija Jain, Samrat Mondal, and
  Aman Chadha.
\newblock A systematic survey of prompt engineering in large language models:
  Techniques and applications.
\newblock {\em arXiv preprint arXiv:2402.07927}, 2024.

\bibitem{houlsby2019parameter}
Neil Houlsby, Andrei Giurgiu, Stanislaw Jastrzebski, Bruna Morrone, Quentin
  De~Laroussilhe, Andrea Gesmundo, Mona Attariyan, and Sylvain Gelly.
\newblock Parameter-efficient transfer learning for nlp.
\newblock In {\em International conference on machine learning}, pages
  2790--2799. PMLR, 2019.

\bibitem{lester2021power}
Brian Lester, Rami Al-Rfou, and Noah Constant.
\newblock The power of scale for parameter-efficient prompt tuning.
\newblock {\em arXiv preprint arXiv:2104.08691}, 2021.

\bibitem{hu2021lora}
Edward~J Hu, Yelong Shen, Phillip Wallis, Zeyuan Allen-Zhu, Yuanzhi Li, Shean
  Wang, Lu~Wang, and Weizhu Chen.
\newblock Lora: Low-rank adaptation of large language models.
\newblock {\em arXiv preprint arXiv:2106.09685}, 2021.

\bibitem{han2024parameter}
Zeyu Han, Chao Gao, Jinyang Liu, Sai~Qian Zhang, et~al.
\newblock Parameter-efficient fine-tuning for large models: A comprehensive
  survey.
\newblock {\em arXiv preprint arXiv:2403.14608}, 2024.

\bibitem{RAISSI2019686}
M.~Raissi, P.~Perdikaris, and G.E. Karniadakis.
\newblock Physics-informed neural networks: A deep learning framework for
  solving forward and inverse problems involving nonlinear partial differential
  equations.
\newblock {\em Journal of Computational Physics}, 378:686--707, 2019.

\end{thebibliography}

\end{document}